\documentclass[AMA,Times1COL]{WileyNJDv5}

\usepackage[inline]{enumitem}

\usepackage{tikz}
\usetikzlibrary{mindmap}

\usepackage{subcaption}

\articletype{Original Article}%

\received{Date Month Year}
\revised{Date Month Year}
\accepted{Date Month Year}
\journal{Journal}
\volume{00}
\copyyear{2024}
\startpage{1}

\newcommand{\joris}[1]{\textbf{\textcolor{cyan}{#1}}}

\begin{document}
\title{Safety Monitoring of Machine Learning Perception Functions: a Survey}

\author[1,2]{Raul Sena Ferreira$^*$}

\author[1,2,3]{Joris Guérin$^*$}

\author[4]{Kevin Delmas}

\author[1,2]{Jérémie Guiochet}

\author[1]{Hélène Waeselynck}

\authormark{Ferreira \textsc{et al.}}

\address[1]{\orgname{LAAS/CNRS}, \orgaddress{\state{Toulouse}, \country{France}}}

\address[2]{\orgname{Université de Toulouse}, \orgaddress{\state{Toulouse}, \country{France}}}

\address[3]{\orgname{Espace-Dev, IRD, Université de Montpellier}, \orgaddress{\state{Montpellier}, \country{France}}}

\address[4]{\orgname{ONERA}, \orgaddress{\state{Toulouse}, \country{France}}}

\corres{*Equal contribution \\ raulsenaferreira@gmail.com\\joris.guerin@ird.fr\\kevin.delmas@onera.fr\\jeremie.guiochet@laas.fr\\helene.waeselynck@laas.fr}

\abstract[Abstract]{Machine Learning (ML) models, such as deep neural networks, are widely applied in autonomous systems to perform complex perception tasks. New dependability challenges arise when ML predictions are used in safety-critical applications, like autonomous cars and surgical robots. Thus, the use of fault tolerance mechanisms, such as safety monitors, is essential to ensure the safe behavior of the system despite the occurrence of faults. This paper presents an extensive literature review on safety monitoring of perception functions using ML in a safety-critical context. In this review, we structure the existing literature to highlight key factors to consider when designing such monitors: threat identification, requirements elicitation, detection of failure, reaction, and evaluation. We also highlight the ongoing challenges associated with safety monitoring and suggest directions for future research.}


\maketitle

\renewcommand\thefootnote{}

\renewcommand\thefootnote{\fnsymbol{footnote}}
\setcounter{footnote}{1}

\section{Introduction}\label{sec:introduction}
Recent advances in Machine Learning (ML) have allowed autonomous systems to leave the safe environment of research labs to perform complex tasks, where failures can have catastrophic consequences. Examples of such safety-critical systems include self-driving cars,\cite{calvi2019runtime} surgical robots,\cite{surgical_robotics} and unmanned aerial vehicles in urban environments.\cite{guerin2021_ssiv} These autonomous systems frequently use large ML models like neural networks for complex sensor signal interpretation, i.e., perception,\cite{premebida2018intelligent} or decision-making, i.e., control.\cite{duan2016benchmarking} This paper focuses on safety mechanisms for critical physical systems relying on ML to process sensor signals.

In various autonomous system applications, essential perception tasks can only be solved using ML. For example, in highly uncontrolled settings, such as self-driving cars\cite{brunetti2018computer} or UAV emergency landing,\cite{guerin2022_icra} deep neural networks must be used to detect pedestrians in RGB images. This information cannot be obtained from other approaches and is crucial to guarantee the system's safety. Despite the great success of modern ML-based perception, it introduces new dependability challenges:\cite{varshney2017safety, faria2018machine, mohseni2019practical}
\begin{enumerate*}
    \item The \textit{lack of well-defined specification}: ML models are learned from examples instead of manually coded, making their operational boundaries elusive, and preventing formal safety guarantees.
    \item The \textit{black-box nature} of the models: traceability and transparency of ML predictions is difficult.
    \item The \textit{high-dimensionality of data}: validation of the complete operational design domain is impossible.
    \item The \textit{over-confidence of neural networks}: output scores cannot be used as is to detect failures since a model can deliver wrong outputs with high confidence.\cite{gal2016dropout}
\end{enumerate*}
Hence, conventional offline safety measures, like fault prevention, removal, and forecasting\cite{avizienis2004basic} are often not sufficient to ensure safety and to certify these systems. Online fault tolerance mechanisms, such as Safety Monitors (SM), emerge as a promising alternative to improve safety in critical systems relying on ML perception. This paper focuses on SMs, which aims to keep the system in an acceptable state during operation, despite faults or adverse scenarios.\cite{machin2018smof} 
Safety monitors are mentioned in the literature under various terms, including safety kernels,\cite{rushby1989kernels} safety managers,\cite{pace2000safety} autonomous safety systems,\cite{roderick2004ranger} checkers,\cite{py2004dependable} guardian agents,\cite{fox2000safe} safety bags,\cite{klein1991safety} or emergency layers.\cite{haddadin2011towards}

\begin{figure}[t]
\centering
\includegraphics[width=0.95\linewidth]{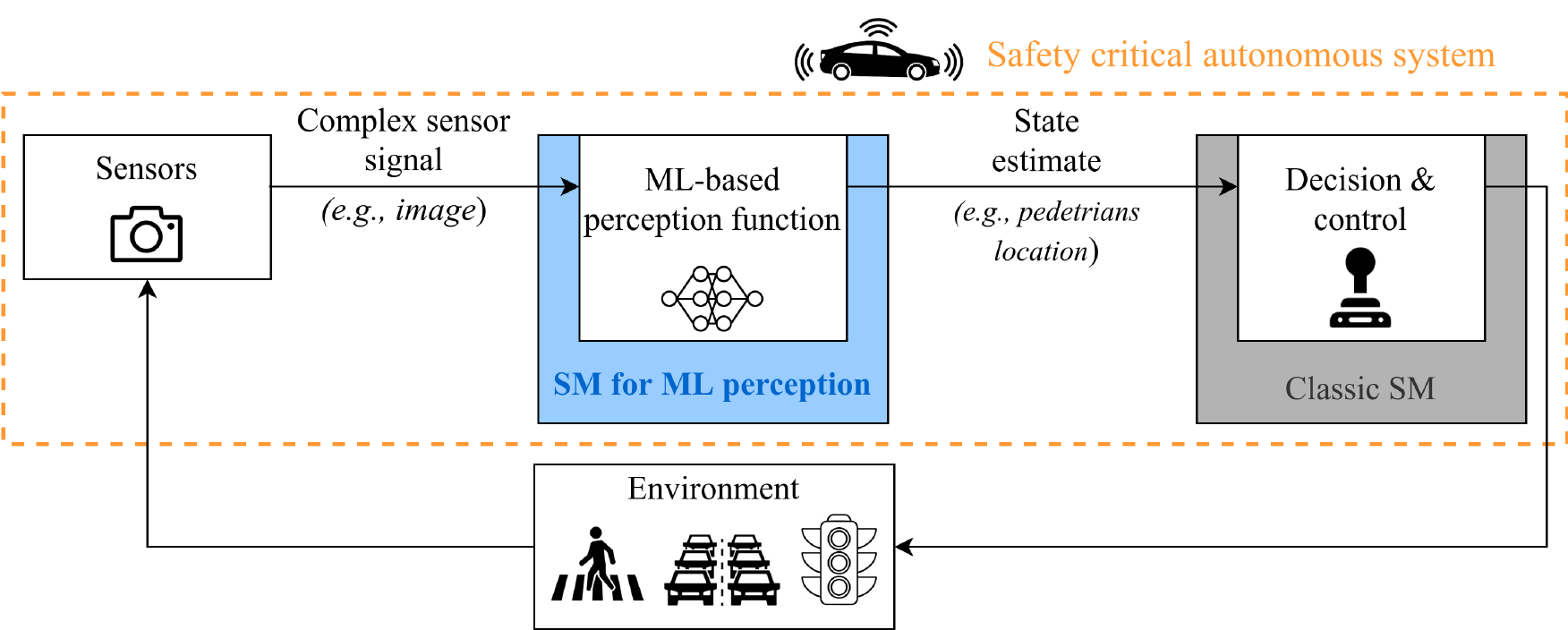}
\caption{\textbf{Safety Monitors for Machine Learning-based Perception Functions.} In modern autonomous systems, state estimation provided by deep learning models cannot be trusted to make safety-critical decisions. Therefore, specific fault tolerance approaches should be implemented to ensure that failures of the ML perception function will not lead to catastrophic outcomes.}
\label{fig:sm_ml}
\end{figure}

Most autonomous systems cannot be considered fault-free given the adverse and unspecified situations they encounter as they evolve in unstructured environments. Fault tolerance approaches aim to ensure that faults do not lead to catastrophic outcomes by designing both error detection and recovery mechanisms. In particular, SMs are components responsible for checking whether specific safety properties are violated and triggering corrective actions. A key characteristic of SMs is their simplicity, to ensure high reliability levels. Traditional SMs monitor system decisions to prevent hazards, but often take for granted the reliability of the state estimation provided by perception components\cite{masson2019safety} (Figure~\ref{fig:sm_ml}). This work considers critical systems relying on ML for perception tasks. 
As explained above, given the inherent complexities of ML-based perception, such outputs should not be trusted blindly to make safety-critical decisions. Therefore, it is essential to design specific safety monitoring approaches for these components, to detect errors early and adapt the system's behavior. 

\begin{figure}[t]
\centering
\includegraphics[width=0.95\linewidth]{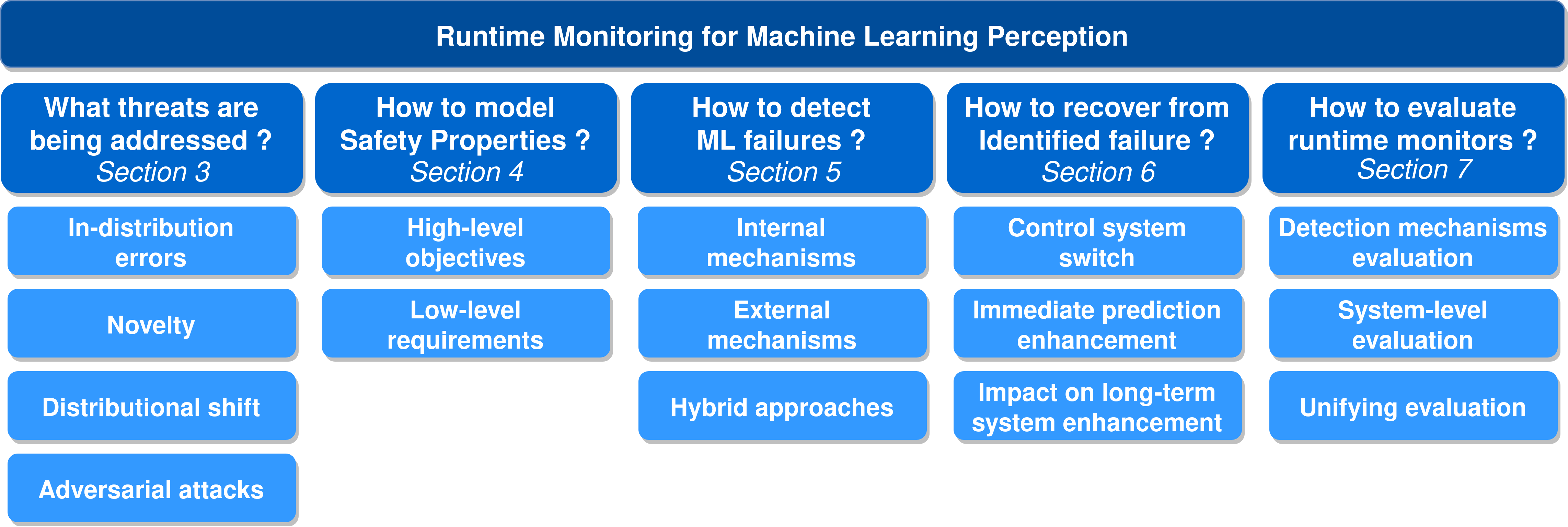}
\caption{\textbf{Key questions to design, implement, and deploy reliable safety monitors for ML perception.} Each section of this work discusses a specific aspect of safety monitoring of machine learning perception.}
\label{fig:taxonomy}
\end{figure}

Many existing studies on SM predominantly concentrate on devising techniques to detect unsafe ML predictions. Consequently, current surveys in this domain focus on categorizing error detection methods, as detailed in Section~\ref{sec:related}. However, SM encompasses more than merely identifying unsafe predictions. In this research, our objective is to comprehensively present the literature pertinent to the creation of robust SMs. As such, our paper is structured around the principal safety considerations intrinsic to SM design:
\begin{itemize}
    \item What threats are being addressed by safety monitors? (Section~\ref{sec:threats}) 
    \item How to derive monitor requirements from safety objectives? (Section~\ref{sec:safety_rules})
    \item Which detection mechanisms can be used? (Section~\ref{sec:detection})
    \item Which recovery actions can be used? (Section~\ref{sec:recovery_actions})
    \item How are safety monitors evaluated? (Section~\ref{sec:evaluation}) 
\end{itemize}
For each of these topics, the main challenges are presented along with existing approaches to tackle them (Figure~\ref{fig:taxonomy}). Although classic SM mechanisms can be useful to ensure the safety of such autonomous systems, this paper focuses on the specificities related to ML perception. As shown in Figure~\ref{fig:sm_ml}, most approaches presented in this paper act close to the ML perception function, often directly considering either its inputs or outputs. However, we cast a broader net, shedding light on diverse strategies pertinent to crafting safety monitors, making this a valuable guide for safety practitioners keen on integrating ML, as well as ML experts navigating the intricacies of safety.

\section{Related Work}
\label{sec:related}

Recent literature about ML safety has flourished, leading to several surveys about the different facets of this domain. This section aims to delineate the contributions of these works and articulate the unique space our study occupies.

In their survey, Xu and Saleh\cite{xu2021machine} categorize the existing research on the use of ML to improve safety across various applications. Similar surveys have been proposed in domain-specific contexts like autonomous driving\cite{muhammad2020deep} and architecture.\cite{hou2021deep} However, these surveys primarily emphasize the benefits ML brings to safety initiatives, sidestepping the crucial aspect of ensuring the safety of the ML components themselves, which is the focus of our study.

The overview proposed by Faria\cite{faria2018machine} is closer to our objective as it presents the challenges posed by ML models in a safety-relevant context and describes potential solutions. However, it does not present the literature for safety monitoring, a critical pillar for ML safety that our survey delves deep into.

Mohseni \etal\cite{mohseni2019practical} also undertook an exhaustive literature review on the safety implications of ML, within the context of autonomous vehicles. Their spectrum of safety mechanisms is composed of three categories: inherently safe design, safety margins, and fail-safe mechanisms, i.e. safety monitors. Regarding SM, they classify detectors by their targeted error types (uncertainty estimation, in-distribution error detection, and out-of-distribution error detection). Likewise, Pandharipande \etal\cite{pandharipande2023sensing} also discusses safety for ML-based automotive perception. For monitoring, they mostly discuss techniques to detect inputs that are outside of the expected operational design domain.

Varshney and Alemzadeh\cite{varshney2017safety} consider ML safety through the lens of risk, epistemic uncertainty, and resultant harm. They discuss four categories of methods to achieve ML safety, including fail-safe strategies, echoing our emphasis on runtime monitoring. Their discussion on monitoring includes classification with rejection and uncertainty estimation techniques. They present several practical examples, but their discussion is mostly focused on the actions that are taken after ML failure has been detected, while not fully covering detection mechanisms.
Recently, Mohseni \etal\cite{mohseni2022taxonomy} also proposed a taxonomy of dependability techniques for ML design and deployment. Starting from engineering safety requirements, they present safety-related ML research into three categories: inherently safe design, enhancing model performance and robustness, and runtime error detection. However, regarding the specific subfield of ML monitoring, which is the focus of our work, they mostly discuss detection mechanisms (uncertainty estimation, outlier detection, and adversarial attacks detection).

Rahman \etal\cite{rahman2021run} journeyed closer to our thematic, presenting a taxonomy for runtime monitoring of ML robotic perception. They organize existing approaches to detect runtime failures into three categories: approaches using past examples of failures to predict future ones, approaches detecting inconsistencies in the perception outputs, and uncertainty estimation approaches. Similarly, Zhang \etal\cite{zhang2023survey} discussed detection approaches, classifying them based on their primary objectives: failure rejection, unknown rejection, and fake rejection.

The surveys presented above are constructed in a bottom-up fashion: they first identify existing works and organize them into relevant categories. This methodology leads to discussions about ML monitoring that are incomplete, as most of the existing literature deals with the detection of unsafe predictions. Traditional safety engineering, however, recognizes error detection as just one component of runtime monitoring and other aspects are also crucial to designing and implementing safe monitors. Therefore, our study was constructed in a top-down fashion, where we first identified primary safety considerations for SM design to frame our survey's structure: threat identification, requirement elicitation, error detection, recovery mechanisms, and performance evaluation. This methodology allows us to go beyond detection mechanisms, highlight areas often overlooked in other surveys, and uncover specific areas where research is lacking from a safety perspective.

Furthermore, unlike other taxonomies that focus on targeted threats, our categorization of detection mechanisms is constructed around how the monitors are integrated within the entire system. We distinguish between internal monitors, which are inherent parts of the ML itself, and external monitors, acting independently from the monitored model. Furthermore, external monitors are classified based on the type of input they analyze -- ML input, ML internal representation, or ML output. We believe that this taxonomy offers greater clarity, especially given the generic nature of many mechanisms that may be deployed against multiple types of threats.\cite{guerin2023outofdistribution}

\section{What threats are being addressed by safety monitors?}\label{sec:threats}

The identification of potential threats is pivotal when designing a dependable SM. Should the safety analysis reveal that a threat is likely to occur for a specific application, this should be reflected in the monitor's evaluation protocol.
Different kinds of threats can affect the ML-based perception functions. Offline threats, such as poor feature engineering,\cite{alasadi2017review} label noise,\cite{bekker2016training} inadequate ML testing,\cite{breck2017ml} or bad model maintenance,\cite{sculley2015hidden} emerge during the ML model development phase. While it is crucial to adhere to robust software engineering practices during SM development, this section mostly focuses on runtime threats, which occur during live operations.

The objective of an SM is to detect unsafe ML predictions and implement corrective actions to prevent catastrophic outcomes. Such hazardous predictions can arise from different types of input data, which we term runtime threats. This section presents a taxonomy of runtime threats for perception functions, inspired by existing literature.\cite{PIM14,chakraborty2018adversarial,granese2021doctor,ferreira2021benchmarking,shen2021towards,ferreira2022simood} Adverse input data are classified as threats if they can be detected or mitigated using similar approaches. Throughout this section, we present runtime threats and discuss their specificities concerning detection, reaction, and evaluation.

Before diving into the different threats, we wish to emphasize that, in our recent work,\cite{guerin2023outofdistribution} we challenged the prevailing understanding of runtime threats in the context of monitoring. A significant portion of NN monitoring research centers on Out-Of-Distribution (OOD) detection, emphasizing the monitor's role in flagging data diverging from the training distribution. However, we highlighted the inherent ambiguity in defining ``OODness'': not all OOD data are inherently dangerous and not all in-distribution data are safe. We advocate for evaluating monitors based on their proficiency in detecting incorrect predictions. Nonetheless, the notion of threats delineated in this section remains crucial for SM design and evaluation. Given that a monitor's efficacy in error detection might fluctuate across diverse threats, it's paramount that every potential threat—identified through a rigorous safety analysis—is considered during the monitor's testing phase.

\subsection{In-distribution errors}

Modern deep-learning architectures have achieved impressive results in many perception tasks. According to the latest leader board on \textit{papers with code}\footnote{https://paperswithcode.com/sota}, the top-performing model for semantic segmentation on Cityscapes\cite{cityscapes} has a mean intersection over the union of $84.5\%$,\cite{segmentation_transformer} the best model for image classification on ImageNet\cite{deng2009imagenet} has a top-1 accuracy close to $91\%$, and the leader for object detection on COCO\cite{coco} has a mean average precision around $63\%$.\cite{liu2021swin} These benchmarks were established on the test splits of these datasets, assumed to derive from the same distribution as the training data, denoted as In-Distribution (ID) data. While these results are excellent and allow researchers to build useful applications, even the best computer vision models are not flawless. To guarantee the system's safety, an SM should be able to handle these errors.

Beyond this fundamental model generalization issue, there is another problem: the data incompleteness. Rare conditions tend to be underrepresented since the training data only account for a small subset of all real-world possibilities.\cite{shafaei2018uncertainty} Such data, presenting different characteristics than the training data, are considered Out-Of-Distribution (OOD)\cite{yang2021generalized} and are discussed as different types of threats in the coming sections. There is no unified naming convention for the threats presented hereafter in the literature, but we strive to propose clear definitions to avoid ambiguity.

\subsection{Novelty threats}
A new input data encountered at runtime is considered ``novel'' when its category/label does not refer to any of the predefined categories known by the model.\cite{yang2021generalized} 
For example, Blum \etal\cite{blum2019fishyscapes} studied the problem of semantic segmentation of a driving scene and trained their model on the Cityscapes dataset.\cite{cityscapes} At runtime, when a dog crosses the road, its corresponding pixels are considered novelty as they do not belong to the predefined set of classes of Cityscapes. Hence, when a novelty input is presented to an ML perception model, it cannot return a correct answer.

The above example shows that facing novelty inputs is common for autonomous systems evolving in unstructured environments. Hence, it is crucial to equip ML perception models with defensive mechanisms against this runtime threat. A typical strategy consists of building classification models with rejection,\cite{condessa2017performance} with the ability to reject uncertain predictions such as objects outside the network scope. Regarding the recovery after detecting novelty threats, the actions implemented 
should not rely on the possibility of obtaining a better estimate of the correct prediction. Concrete approaches to detect novelty threats, recover from them, and evaluate the ability of an SM to address them are discussed in Sections~\ref{sec:detection}, \ref{sec:recovery_actions}, and Section~\ref{sec:evaluation}, respectively.

\subsection{Distributional shift threats}
A distributional shift occurs when the marginal distribution of the runtime input data is different from the training distribution, while the label generation mechanism remains unchanged.\cite{shen2021towards} Regarding safety monitoring, we distinguish two types of shifts: covariate and semantic. 

\subsubsection{Covariate shift}
A covariate shift is a condition that decreases the ML performance through time in dynamic environments.\cite{ferreira2019amanda} In other words, covariate shift threats are new data presenting different characteristics in their composition but for which the semantic content is not different from training. For images, such threats are also called corruptions or perturbations and were presented and discussed extensively by Hendricks and Dietterich.\cite{hendrycks2019benchmarking} These deteriorated data can come from:
\begin{itemize}
\item \emph{Failure in exteroceptive sensors}. These perturbations come from hardware defects and include various errors such as pixel traps, shifted pixels, and Gaussian noise. There are specific approaches to identify sensor faults,\cite{khalastchi2013sensor} which can be addressed by tuning the sensor parameters\cite{micheloni2009active} or having a backup sensor system.\cite{surya2018deployment}

\item \emph{Changes in external conditions}.  For autonomous systems evolving in unstructured environments, the training data cannot cover all possible real-world conditions. For example, outdoor perception functions should work for different kinds of weather (e.g., snow, fog) and lighting conditions (e.g., night, sunset). As illustrated in the disengagement reports by major companies, such external factors influence the perception performance and can reduce the safety of autonomous vehicles.\cite{sinha2021crash} 
\end{itemize}

To deal with these two covariate shift types, both traditional signal processing\cite{motwani2004survey} and modern deep learning approaches\cite{tian2020deep} have been used to detect and reduce data noise. However, the techniques used against covariate shift threats depend highly on the amount of noise in the data.

\subsubsection{Semantic shift}
A semantic shift threat refers to images showcasing objects that:
\begin{itemize}
    \item presents different attributes than known members of this category, such as a pedestrian detector trained in summer encountering individuals wearing winter clothes,\cite{rasouli2018s}
    \item displays uncommon interactions between known classes and their surroundings, like an overturned truck.\cite{stumpf_2020}
\end{itemize}
In contrast to novelty threats, semantic shift only includes data where the objects align with the model's predefined categories. While one might see a semantic shift as a particular case of covariate shift, its distinct challenges in safety monitoring set it apart. Notably, semantic shifts cannot be handled with denoising or backup sensors. Instead, some studies have explored the detection of object attributes that remain consistent across varying environments.\cite{zhang2020attribute} For instance, when confronting a pedestrian detector with shifts in clothing attributes, detecting faces rather than full bodies could be more effective. Though not designed for ML monitoring, such approaches hold promise for identifying model failures linked to semantic shift threats.

\subsection{Adversarial threats}\label{sec:adversarial}
An adversarial input is an intentional modification of in-distribution data to make ML models fail with high confidence.\cite{akhtar2018threat, kurakin2016adversarial}
In real-world scenarios, these malicious attacks can be made by applying modifications on targeted physical objects such as painting black lines on the road to force the ML model to interpret it as a road lane.\cite{boloor2020attacking}

Adversarial threats can lead to serious safety issues if applied against the perception functions of critical systems. Therefore, they should be handled by specific SMs as they are likely to fool generic monitoring approaches. However, specific hardening approaches, such as gradient hiding and defensive distillation, have been developed to identify them or increase model robustness.\cite{chakraborty2018adversarial}

\section{How to derive safety monitors from safety objectives?}\label{sec:safety_rules}

Safety monitoring guarantees that some safety properties are not violated, despite potential faults occurring in the main system. The elicitation and modeling of these properties are essential steps in designing safety monitors. For instance, Machin \etal\cite{machin2018smof} used a HAZOP-UML hazard analysis\cite{guiochet2010experience} to identify high-level safety objectives expressed in natural language. These high-level objectives are then converted to low-level safety requirements, expressed formally in the system's state space, and observable by the monitor. For a mobile robotic platform in a standard industrial setting, an example of a high-level safety objective is ``the robot platform must not collide with a human". A low-level safety requirement can be derived by comparing the braking distance with the distance of any obstacle sensed by a laser. In this example, the low-level requirements are easy to express and implement since the sensor signal can be interpreted in terms of the high-level requirement.

The high-level safety objectives can still be identified using standard hazard analysis tools for complex systems involving machine learning perception. However, converting them into low-level monitoring requirements is not straightforward. Indeed, expressing and implementing a high-level requirement in raw sensors can result in solutions that are too conservative or even infeasible to be deployed at runtime.
For example, if we consider an emergency braking system (EBS) implemented in an autonomous vehicle:
\begin{itemize}
    \item Using \emph{simple sensor signals} such as a laser is not enough to capture the semantic information required to distinguish between pedestrians and other moving vehicles. 
    Such semantic information is crucial for EBS to perform two very different low-level requirements: to stop the ego vehicle when the EBS identifies an object as a pedestrian, or slightly decelerate the ego vehicle when the EBS identifies an object as a moving vehicle. 
    Therefore, stopping the car for all sensed objects is \emph{too conservative}, which would significantly alter the availability of the system.
    
    \item Using \textit{complex sensor signals} such as RGB image pixels from camera sensors is not enough to guarantee that a high-level objective is not violated.
    That is, measuring the pixels alone is \emph{infeasible} to perform the EBS task since such raw RGB values cannot give insightful information for the EBS to perform a high-level requirement such as avoiding a collision.
\end{itemize}
Hence, we should specifically monitor the ML function responsible for localizing pedestrians. In other words, the system-level safety objectives should be expressed as variables related to the ML model (input, activation, output).

As explained above, most current works on ML monitoring focus on detecting when a model is wrong and should not be trusted. This is a good generic formulation of the problem, agnostic of the system in which the model is embedded. However, we believe that using information from the application context to refine the low-level monitor requirements is a promising research direction. In particular, the hazard analysis of the system could be used to identify safety-critical regions of the ML model input/output space or to understand under which system configuration an ML error is hazardous. In addition, building monitors for specific sub-regions of the state space might allow us to come up with more effective local monitors and better allocation of resources. 

Although this lead has not yet been explored for ML monitoring, some research from ML safety could serve as a first step towards building better specific monitors. In their work, Dreossi \etal\cite{dreossi2019compositional, dreossi2019verifai} propose to identify regions in the state space where a failure of the ML model results in a violation of a formal specification. For an autonomous vehicle use case, they show that errors of an ML-based obstacle detection model are only threats for certain state configurations (speed and distance to other vehicles). On the other hand, Salay \etal\cite{salay2019safety} introduced an approach called Classification Failure Mode Effects Analysis (CFMEA) to study the safety of an ML classifier. It serves to identify the kind of errors that can lead to a safety-critical situation. For example, CFMEA can assess the severity of different control actions based on different classification errors in an autonomous vehicle scenario. This approach represents a promising research direction for runtime monitoring of ML perception functions. For example, knowing that an ML failure would only cause catastrophic events in some subsets of the state space could help to collect better data to design monitors in these specific regions.

\section{Which detection mechanisms can be used for safety monitoring?}\label{sec:detection}

Traditional monitors analyze both exteroceptive (e.g., distance) and proprioceptive (e.g., speed) sensors to detect safety threats. 
They apply simple rules on sensor data based on formal specifications.\cite{machin2018smof} However, for complex perception functions, these monitors struggle to interpret raw sensor signals like image pixels. This section discusses recent strategies to detect errors in ML model signals. Yet, even in ML-augmented autonomous systems, traditional monitoring remains indispensable to handle other sensors and ensure the system's proper functioning.

Despite its significance, the specific field of ML safety monitoring has received limited research attention. Nonetheless, various ML techniques, hailing from sub-fields like uncertainty estimation, anomaly detection, ensemble methods, or multi-modal perception, show potential as SM detection mechanisms. This section delves into such prospective approaches, encompassing those not specifically designed for safety monitors.

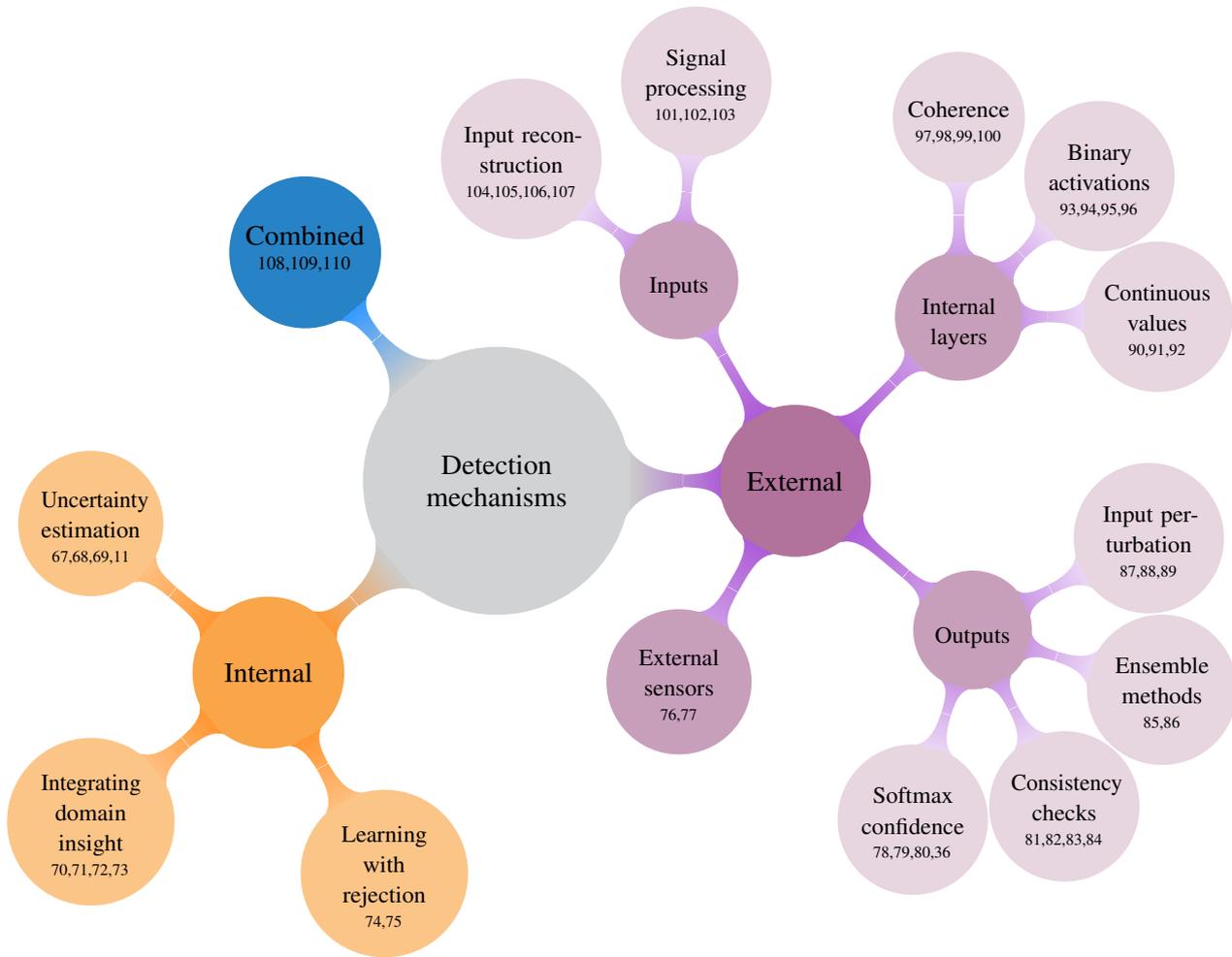
\begin{figure*}
\centering

\resizebox{0.95\linewidth}{!}{
\begin{tikzpicture}[mindmap, grow cyclic, 
                    every node/.style={concept}, 
                    concept color=Gray!40, 
                    level 1/.append style={level distance=4.5cm,sibling angle=140, font=\large},
	                level 2/.append style={level distance=3.5cm,sibling angle=80, font=\normalsize},
	                level 3/.append style={level distance=3cm,sibling angle=45, font=\normalsize, text width=1.6cm}]
\node{Detection mechanisms}
child [concept color=BurntOrange!80] { node {Internal}
    child [concept color=BurntOrange!50] { node {Uncertainty estimation\\ \cite{huang2018efficient, hubschneider2019calibrating,  costante2020uncertainty, gal2016dropout}}}
    child [concept color=BurntOrange!50] { node {Integrating domain insight\\  \cite{chhablani2021superpixel, ramanathan2015learning, xu2018semantic, donadello2017logic}}}
    child [concept color=BurntOrange!50] { node {Learning with rejection\\  \cite{cortes2016boosting, geifman2019selectivenet}}}
}
child [concept color=DarkOrchid!80] { node {External}
    child [concept color=DarkOrchid!50] { node {External sensors\\ \cite{zhou2019automated, ramanagopal2018failing}}}
    child [concept color=DarkOrchid!50] { node {Outputs}
        child [concept color=DarkOrchid!20] {node {Softmax confidence\\ \cite{hendrycks2016baseline, liang2018enhancing, hsu2020generalized, granese2021doctor}}}
        child [concept color=DarkOrchid!20] {node {Consistency checks \\ \cite{kang2018model, harper2021safety, chen2021monitoring, guerin2020robust}}}
        child [concept color=DarkOrchid!20] {node {Ensemble methods\\ \cite{yahaya2019consensus, roitberg2018informed}}}
        child [concept color=DarkOrchid!20] {node {Input perturbation\\ \cite{kantaros2020visionguard, wang2019adversarial, liu2020input}}}
    }
    child [sibling angle=90, concept color=DarkOrchid!50] { node {Internal layers}
        child [concept color=DarkOrchid!20] {node {Continuous values \\ \cite{rahman2019did, sun2021react, lukina2020into}}}
        child [concept color=DarkOrchid!20] {node {Binary activations\\ \cite{cheng2019runtime, henzinger2020outside, ferreira2023sena, wu2021customizable}}}
        child [concept color=DarkOrchid!20] {node {Coherence\\ \cite{wang2020dissector, schorn2020facer, lee2018simple, chen2020task}}}
    }
    child [concept color=DarkOrchid!50] { node {Inputs}
        child [sibling angle=70, concept color=DarkOrchid!20] { node {Signal processing\\ \cite{ndong2011signal, kim2006image, liao2013video}}}
        child [concept color=DarkOrchid!20, text width=1.8cm] { node {Input reconstruction\\ \cite{sabokrou2018adversarially, denouden2018improving, stocco2020misbehaviour, cai2020real}}}
    }
}
child [sibling angle=130, concept color=RoyalBlue!80] { node {Combined\\ \cite{loquercio2020general, cofer2020run, buerkle2021fault}}
};
\end{tikzpicture}
}
\caption{\textbf{Taxonomy of detection mechanisms}. A visual representation of the different types of approaches to detect a failure of a critical ML-based perception function.}
\label{fig:detection_mechanisms}
\end{figure*}

This section presents a comprehensive taxonomy of detection mechanisms. Our categorization of detectors revolves around the manner in which they are assimilated into the complete system (Figure~\ref{fig:detection_mechanisms}). Within each category, we outline the primary approaches and assess their advantages and limitations. However, it is important to mention that current research does not allow us to reach definitive conclusions regarding the efficacy of these techniques. Indeed, the literature currently presents a varied landscape with differing evaluation methods and conflicting experimental results. Notably, a study by Ferreira \etal\cite{ferreira2021benchmarking} revealed unsatisfactory results for several evaluated methods, despite the good results presented in the original papers.

\subsection{Internal mechanisms}
Internal detection mechanisms are approaches where the ML model itself is trained to predict its failures. In other words, the NN is designed to return both predictions and information regarding the trust in these predictions. We classify internal mechanisms into three families of approaches.

\subsubsection{Uncertainty estimation}
Uncertainty estimation in deep learning has been widely studied recently.\cite{mena2021survey, gawlikowski2021survey, ulmer2021survey} Most deep learning models produce a single output value per input data. Uncertainty estimation approaches replace point estimate predictions with a probability distribution over the output space. These probabilities can then be used to evaluate the risk of trusting the prediction. For example, for deep learning classifiers, the outputs of the softmax layer define a probability distribution over the possible classes. However, the community has widely questioned using raw softmax output as an uncertainty proxy. Softmax is merely a normalization technique, not designed to represent meaningful probabilities, and thus often yields overconfident predictions.\cite{sensoy2018evidential, hendrycks2016baseline} 

In Bayesian Deep Learning, the weights of an NN are treated as random variables, and the objective is to learn their distributions from the training data. Then, using the Bayes rule, one can compute the distribution of the predictions. Traditional Bayesian statistics\cite{seeger2006bayesian, box2011bayesian} is a natural way to reason about uncertainty in predictive models but comes with a prohibitive computational cost to be used in practice. Recently, various approaches have been proposed to compute approximate Bayesian inference on large ML models, including Variational Inference,\cite{chen2018variational, milios2018dirichlet, malinin2018predictive, rossi2019good} Laplace approximation\cite{lee2018simple} and sampling methods.\cite{welling2011bayesian} For a detailed review of Bayesian deep learning, we refer the reader to the following works.\cite{goan2020bayesian, mena2021survey} 

Bayesian deep learning has been applied to various perception tasks related to autonomous systems, including object detection,\cite{feng2021review} semantic segmentation,\cite{mukhoti2018evaluating, huang2018efficient} end-to-end vehicle control\cite{hubschneider2019calibrating} or visual odometry.\cite{costante2020uncertainty} Of the techniques available, those based on Monte-Carlo Dropout\cite{gal2016dropout, gal2017concrete} are particularly popular due to their simplicity of implementation. This approach retains Dropout layers during inference, enabling stochastic predictions. By repeatedly running the model and analyzing prediction statistics, it evaluates uncertainty.

In theory, a clear understanding of model uncertainty is truly all that is required for constructing robust SMs for ML models. However, methods that provide provably accurate uncertainty estimates are intractable for real-time tasks in autonomous systems. To circumvent this challenge, various approximation methods have been introduced. Depending on their level of simplification, they either remain too computationally expensive or fail to offer reliable uncertainty estimates suitable for safety-critical systems. However, given the rapid advancements in this research area, the landscape could change soon.

\subsubsection{Incorporating domain knowledge}
Domain knowledge can be leveraged to improve the training of ML models. It can be incorporated into the architecture and training process in the form of logical or numerical constraints.\cite{dash2021incorporating} For example, a pioneering work proposed to build an object detection model by training a hierarchy of classifiers using lexical-semantic networks to represent prior knowledge about inter-class relationships.\cite{marszalek2007semantic} This architecture can be used to detect anomalies in the runtime predictions. Likewise, information about the relationship among different superpixels of an image is used in\cite{chhablani2021superpixel} to build a robust classification pipeline. The superpixel relationships are modeled using a graph neural network, which processes the image jointly with a convolutional neural network in the final architecture. This additional information in the model itself can be used to detect incoherence in the final predictions. Ramanathan \etal\cite{ramanathan2015learning} builds an action retrieval model by incorporating other small linguistic, visual, and logical consistency-based actions to effectively identify relationships between unobserved actions from known ones. Other such approaches include attempts to incorporate symbolic knowledge,\cite{xu2018semantic} as well as first-order fuzzy logic to reason about logical formulas describing general properties of the data.\cite{donadello2017logic} 

Such domain-aware approaches are notably effective as they are specifically tailored to monitor relevant elements to a particular application. However, their limitation lies in their lack of universality. Each application or task requires a unique design, demanding considerable expert effort for every subsequent model. At times, crafting them to fit certain tasks might even be unattainable. Moreover, such niche approaches seem to have a higher rate of false positives, as indicated by Ferreira \etal\cite{ferreira2021benchmarking}

\subsubsection{Learning with rejection}
In the setup of selective classification -- also called classification with rejection -- input data can either be classified among one of the predefined categories or be rejected, i.e., the system produces no prediction. Such rejection mechanisms can actually be viewed as built-in safety monitors as they allow to reject uncertain predictions at runtime.
This kind of approach has been presented as a promising way to control the confidence in the monitored model in critical autonomous driving scenarios.\cite{mohseni2019practical} These approaches are internal mechanisms as they consist of modifying the model and learning algorithm to account for rejection. In other words, the predictor and the rejection function are trained jointly and are part of a single unified model. Several approaches have been proposed to integrate rejection options to traditional ML models such as support vector machine,\cite{fumera2002support} K-nearest neighbors\cite{hellman1970nearest}, and boosting.\cite{cortes2016boosting} Recently, Geifman and El-Yaniv\cite{geifman2019selectivenet} presented Selectivenet, a neural network architecture optimized to perform 
classification and rejection simultaneously. Several other approaches have been introduced recently for learning with rejection, as discussed in more detail in the recent survey by Hendrickx \etal\cite{hendrickx2021machine}

Utilizing an integrated model for prediction and rejection has its advantages. When optimized together, these elements can boost performance for specific tasks, potentially reducing inherent biases due to synergistic training. However, the drawback of such integrated techniques is that any alteration to the monitor requires a complete model retraining, further taxing computational resources. We also believe that learning with rejection does not align well with today's machine learning landscape, where leveraging open-source models as a foundation is common. Indeed, such an approach prevents seamlessly integrating a monitor without the labor-intensive need for total retraining. Finally, while such unified models might diminish bias, there is a counter-argument in favor of diversifying the design. By separating the prediction and monitoring processes, a blend of expertise -- spanning data scientists to safety experts -- can collaborate, fostering a multi-faceted, bias-resistant system. Designing a good rejection-enabled model requires specific expertise beyond basic machine learning knowledge, which might not always be easy to find.

\subsection{External mechanisms}

External detection mechanisms are independent components in charge of monitoring the behavior of an ML model during execution. As they are not directly tied to the ML model, they are not required to be trained jointly and can be developed later by specific safety teams. We identified different mechanisms in the literature that differ in their position in the ML perception pipeline. In particular, external detection mechanisms can monitor either the ML model inputs, internal representations, outputs, or even data from other sources.

\subsubsection{Monitoring the DNN inputs} 
Some approaches predict failures of an ML perception model by monitoring its inputs, e.g., the raw images. These approaches are independent of the monitored ML model, as they characterize the expected operational conditions under which a neural network can be used and discard new abnormal input data before the perception function processes them. Next, we present the existing approaches.

\paragraph{Traditional approaches} 
Traditional signal-processing approaches can be used to identify an anomalous sensory input before it enters the ML model.\cite{ndong2011signal} These approaches characterize some statistical patterns of ``normal'' data (i.e., from the training set) and compare them to new online inputs. In particular, for images, one can identify noise patterns of the camera and standard lighting conditions and detect abnormal images using standard image processing.\cite{kim2006image} This kind of approach has been used to identify water droplets in images for autonomous driving scenarios.\cite{liao2013video} 

These approaches are good at detecting sensor defects and variations in external conditions like lighting and weather. Yet, they fall short of identifying subtle changes in complex signals, such as novelty classes or semantic shifts.

\paragraph{Input reconstruction} 
Recent techniques have relied on unsupervised deep learning to identify anomalous images. They start by training an auto-encoder that learns a lower-dimensional latent representation of the images and how to reconstruct the original images from it. Then, at runtime, the auto-encoder can be used to decide if a new image is an outlier by comparing its reconstruction error to a fixed threshold defined during training.\cite{sabokrou2018adversarially, denouden2018improving, stocco2020misbehaviour,cai2020detecting} Another approach consists of training an outlier detection model on the latent representation of the auto-encoder to predict the nonconformity of new inputs.\cite{cai2020real} In practice, this family of approaches was used to identify unexpected conditions, such as weather changes, to anticipate the misbehavior of an autonomous vehicle within a simulation environment.\cite{stocco2020towards} Finally, Feng and Easwaran\cite{feng2021robust} proposed to detect unusual movements in real-time by combining optical flow operation with representation learning via a variational auto-encoder. 

Compared to conventional techniques, these methods have a higher likelihood of detecting semantic shifts in data, given their reliance on deep representation learning from the training dataset. However, two neural networks trained on the same dataset may not necessarily learn identical features.\cite{guerin2021combining} Relying on such an auto-encoder could introduce an additional bias if its generalization diverges from that of the model being monitored.

\paragraph{Introspection}
A well-explored technique in the robotics domain aims at predicting future failures at runtime. For instance, Gurau \etal\cite{guruau2018learn} proposed two models that predict perception performance from observations gathered over time. Then, the monitor can switch control to a human operator if the robot's perception system is predicted to underperform. Kuhn \etal\cite{kuhn2020introspective} proposed an introspective approach to predict future disengagements of the car by learning from previous disengagement sequences. They monitor both input images and other state data from the car.

Relative to other input monitoring methods, these techniques stand out as they strive to assimilate the underlying knowledge of the monitored ML. They learn a fresh data representation to identify unsafe input under the supervision of the model itself. However, it appears that these methods could be enhanced by accessing additional details about the monitored model, like its activations, which might aid in devising superior models for detecting unsafe input data.

\paragraph{Insights regarding input monitoring}

Input monitoring techniques present the advantage of being independent of the monitored neural network, which facilitates the software engineering process. Yet, this very independence can be a double-edged sword, making it challenging to anticipate ML failures on specific inputs without direct model inspection. Such techniques are effective when the operational design domain (ODD) is clearly defined and the predictor is reliably error-free within its ODD. Otherwise, employing techniques tailored to the specific predictor is advantageous, as they are more likely to capture the model's true strengths and vulnerabilities.

\subsubsection{Monitoring the DNN internal representations}
Other detection approaches monitor the values from the ML model's hidden layers. The underlying principle is that the training data alone does not fully encapsulate the model's understanding and that crucial information is embedded within the model's weights. This can be justified intuitively by the fact that different models behave differently with new inputs, even when trained on the same dataset.\cite{gontijo2021no, guerin2021combining} This section discusses approaches to monitor neural network activations at runtime.

\paragraph{Continuous layer values} 
Several works have proposed to detect unreliable predictions by analyzing the output values of certain layers for novel input data. Rahman \etal\cite{rahman2019did} trained a binary anomaly classifier on features extracted from a hidden layer of a neural network. Another recent work proposed truncating feature activations from the penultimate layer, before the classification head, to get better uncertainty estimates.\cite{sun2021react} Adopting a different methodology, Lukina \etal\cite{lukina2020into} employed a centroid-based clustering technique on the internal representations of a designated layer to characterize known inputs, leveraging the distance to cluster centers as an indicator to filter out atypical data at runtime. Finally, Wang \etal\cite{wang2021real} proposed to monitor the neurons within a faster R-CNN by representing distributions of activation patterns and calculating the Kullback-Leibler divergence between them.

The raw values of internal layers activation contain rich information about the data being processed. Initial layers capture raw data intricacies, while subsequent layers focus on data interpretation. Yet, the volume of features extracted from these layers can be extensive, leading to significant computational demands, which might be unfeasible for some constrained real-time applications.
        
\paragraph{Binary layer activations}
To reduce the memory usage of internal representation monitors, other works have proposed looking only at the binary activations of a given layer. As ReLU is one of the most popular activation functions in deep neural networks, one can inspect whether a new input is triggering an activation of a specific neuron or not, i.e., non-zero value. The advantage of considering such binary variables is that they can be stored easily using binary decision diagrams\cite{cheng2019runtime} or abstraction boxes.\cite{henzinger2020outside, ferreira2023sena, wu2021customizable} Then, abnormal data are identified by comparing activation patterns encountered at runtime to those recorded during training.

\paragraph{Coherence between layers} 
Several works have explored the potential of simultaneously examining multiple hidden layers. Wang \etal\cite{wang2020dissector} introduced Dissector, a tool designed to ascertain if the outputs across different layers yield consistent decisions. Schorn and Gauerhof\cite{schorn2020facer} proposed an approach called FACER, which builds a feature vector capturing activations from various layers by summing the values of each feature map. Similarly, Lee \etal\cite{lee2018simple} fitted class conditional Gaussian distributions to both low-level and upper-level features of the deep learning model and defined a confidence score based on the Mahalanobis distance. Another innovative approach consists of tracking the model's internal representations backward to build a saliency map of a given input.\cite{chen2020task} The patterns of this map are then compared with the ones obtained for the training set within the same category. Recently, Wang et al.\cite{wang2022vim} proposed ViM (Virtual-logit Matching), combining a feature-based class-agnostic score and a logits-based class-dependent score, which obtained very good results for detecting out-of-distribution data.

Leveraging multiple hidden layers often proves advantageous, primarily because it alleviates the challenge associated with optimal monitoring layer selection. Yet, this approach compounds the issues related to computational time. As such, striking a balance between the number of layers to incorporate and the system's inherent constraints becomes pivotal.

\paragraph{Insights regarding internal layers monitoring}
The activations within internal layers provide an invaluable insight into a model's comprehension of the data it processes, making them pivotal for crafting effective monitors. However, the challenge lies in pinpointing the right layer for monitoring to ensure optimal outcomes – a decision far from straightforward. Marrying the ideal layer with the most suitable monitoring transformation is equally vital. Owing to these intricate selections, current results showcase significant variance. Different methodologies and layers excel for varying datasets, but the community has yet to converge on definitive guidelines on which techniques and layers best serve specific scenarios. Consequently, formulating an effective internal layer monitor remains a challenging endeavor.

\subsubsection{Monitoring the DL outputs}\label{sec:detection_outputs}
Some detection approaches monitor the ML outputs. For example, for classification tasks, the output contains information about the target class and the model's confidence associated with this label.

\paragraph{Manipulation of softmax confidence} 
A straightforward strategy to monitor deep neural network outputs consists of properly establishing what values of the (softmax) confidence score can be considered reliable.\cite{hendrycks2016baseline} Several enhancements over this baseline have been proposed to address the calibration issues from softmax confidence scores. Liang \etal\cite{liang2018enhancing} presented ODIN, which uses temperature scaling and small input perturbations to separate the softmax scores between in- and out-of-distribution data. Hsu \etal\cite{hsu2020generalized} modified this approach to function without requiring out-of-distribution examples and further enhanced detection capabilities using confidence score decomposition. A more recent methodology, DOCTOR, aimed to characterize the optimal discriminator, focusing solely on the softmax probability.\cite{granese2021doctor} Finally, Liu \etal\cite{liu2020energy} questioned softmax's efficacy for reliable uncertainty estimation, suggesting an alternative energy score for logit transformation.

The main advantage of these techniques lies in their straightforward implementation, only requiring attention to the neural network's outputs, and ensuring minimal computational and memory burden. They have shown promise in detecting threats such as novelty, covariate shifts, and adversarial attacks. However, their performance appears to drop for the task of detecting actual model errors. It's also worth noting that these strategies primarily cater to monitoring classification models.
    
\paragraph{Consistency checking}
Several approaches focus on verifying the spatial or temporal consistency of a sequence of predictions. One primary method involves employing expert knowledge to establish constraints on output sequences. For instance, Kang \etal\cite{kang2018model} built a monitor for object detection by identifying flickering, i.e., an object should not keep appearing and disappearing in successive frames of a video. Another strategy, showcased by Harper \etal\cite{harper2021safety}, translates the official highway code's rules and clauses into logical assertions for monitoring. 

Alternatively, these patterns can be learned from data. Chen \etal\cite{chen2021monitoring} proposed a logical framework to evaluate both temporal and spatial coherence of bounding box predictions to identify erroneous detections.
Temporal coherence focuses on how bounding box labels evolve across sequential frames, while spatial coherence learns standard bounding box sizes at various locations. On another note, Guerin \etal\cite{guerin2020robust} proposed a consistency monitor tailored for objects under periodic motion, like those on production lines. They train a Gaussian process to estimate the probability for a bounding box to be at a particular location at a specific time and use it to discard erroneous detections. Finally, Rabiee and Biswas\cite{rabiee2019ivoa} detect failures of stereo vision-based perception through inconsistencies in plans generated by a vision and a supervisory sensor.

These techniques are primarily tailored for object detection tasks. One of the advantages of these methods is the potential to gather samples for subsequent ML training, either through human annotation or weak supervision.

\paragraph{Ensemble methods} 
Using an ensemble of ML models is a conventional approach to enhance robustness. An effective ensemble should consist of models that are ``good'', ``independent'' and ``sufficiently numerous''. 
For complex ML used in perception tasks, such ensembles can be composed of models with the same architecture but trained to identify different aspects of the data or with different architectures trained with the same data.\cite{theodoridis2015machine} For deep neural networks, Gontijo \etal\cite{gontijo2021no} conducted a study about the influence of different training parameters on what the model knows, which can be used to maximize the ensembling benefits.

While neural network ensembles are commonly used to enhance the robustness of ML systems, they can also serve as effective tools for monitoring predictions. Ensembles provide additional information, such as the level of agreement among individual components, which can be valuable for assessing an ML system's confidence in its predictions. When computational resources are not a limiting factor, building ensembles of deep neural networks presents a promising approach to monitor the coherence between individual models and mitigate the propagation of errors from a single neural network.

In practice, different voting strategies were proposed to address anomaly detection. Yahaya \etal\cite{yahaya2019consensus} weight each model vote based on its performance and a score for what is considered normal. Roitberg \etal\cite{roitberg2018informed} leverage the estimated uncertainty of each prediction to measure novelty. Roy \etal\cite{roy2022runtime} present a runtime monitor based on predictive processing and dual-process theory.
They developed a bottom-up neural network comprising two layers:
\begin{enumerate*}
    \item a feature density model that learns the joint distribution of the original inputs, outputs, and the model’s explanation for its decisions. 
    \item a graph Markov neural network that captures an even broader context.
\end{enumerate*}

\paragraph{Robustness to input perturbation} 
To verify if a new input can be considered safe, it was proposed to measure its sensitivity to input perturbations. Such perturbations can be applied either to the data (e.g., image compression\cite{kantaros2020visionguard}) or to the ML model itself through random mutations.\cite{wang2019adversarial} Another possibility is to check the stability of a model within a radius distance calibrated during the training.\cite{liu2020input} The underlying hypothesis for these approaches is that, for valid input, the outputs of the neural network should be robust to small perturbations. 

This approach is mostly used to detect adversarial inputs, but it has also been used for practical scenarios involving critical systems. For example, Zhang \etal\cite{zhang2018deeproad} proposed DeepRoad, a GAN-based metamorphic testing and input validation framework for autonomous driving systems. 

Similarly to ensemble methods, a primary limitation of these techniques is their need for multiple neural network forward passes for each input. This can lead to extended processing times or substantial computing resource demands, which may not suit many autonomous systems, especially those with real-time and power efficiency constraints.

\paragraph{Insights about output monitoring}

Observing the neural network outputs has the inherent benefit of directly reflecting the essential information relayed to the system's decision module. In contrast, the output layer, compared to early internal representations, contains less granular details, necessitating the incorporation of additional context for optimal detection outcomes.

\subsubsection{External sensors}
To conclude this section, we present one last family of approaches to detect DNN failures. It aims at checking whether the ML predictions are consistent with signals coming from other sensors. This way, Zhou \etal\cite{zhou2019automated} proposed to use an additional LIDAR sensor to monitor the runtime behavior of a semantic segmentation model. By checking the consistency of geometric properties between the predicted segmentation map and the LIDAR points, they can measure the segmentation model accuracy at runtime. Similarly, Ramanagopal \etal\cite{ramanagopal2018failing} used a second camera to monitor the results of an object detection model. Inconsistencies in the object detector outputs between a pair of similar images are used as a hypothesis to detect false negatives, i.e., missed detections.
Finally, Li \etal\cite{li2022camera} present an approach to monitor extrinsic camera calibration quality by using inertial measurement unit (IMU) data to capture mismatches of road image features. 

These techniques are very powerful in detecting inconsistencies in ML predictions, bolstering the safety of autonomous systems with ML-driven perception.  However, their effectiveness is primarily confined to ML tasks with a geometrical component, like segmentation or detection. For purely semantic tasks, such as classification, simple sensor signals cannot be used to provide adequate monitoring.

\subsection{Combined detection approaches}
Various approaches have combined monitors to build more robust detection systems. Loquercio \etal\cite{loquercio2020general} proposed to monitor data and model uncertainty using distinct mechanisms. Data uncertainty is assessed by propagating sensor noise characteristics through the network, while model uncertainty is assessed using Monte-Carlo Dropout,\cite{gal2016dropout}. The cumulative uncertainty is obtained by combining both sources using stochastic Assumed Density Filtering. Buerkle \etal\cite{buerkle2021fault} presented an approach using two types of detection mechanisms called sensor checks (monitoring model input with an auto-encoder) and plausibility checks (monitoring spatiotemporal coherence of model predictions). Meanwhile, Cofer et al.\cite{cofer2020run} integrated four distinct monitors for end-to-end aircraft taxiing, harmonizing data from various sensors, standard computer vision algorithms, and input reconstruction approaches to forge a resilient neural network monitoring mechanism. Lastly, Guerin et al.\cite{guerin2022_icra} combined three monitors for drone emergency landings: Monte-Carlo Dropout, local resolution enhancement, and classification hierarchy.

Throughout this section, we've delved into a myriad of recent techniques for detecting unsafe predictions. These methods each bring unique qualities to the table, addressing various data types, tasks, implementation nuances, and integration methodologies. While the individual efficacy of these techniques in the realm of safe autonomous systems is still under examination, their collective use holds substantial promise in bolstering safety and paving the way for system certification. However, pinpointing the optimal amalgamation of these strategies remains a challenge, necessitating a profound understanding of prevailing methodologies, proper identification of potential threats through safety analysis, and a good understanding of system constraints and requirements. Through this survey, we aspire to assist in this complex endeavor by offering safety experts a complete view of the existing methodologies and illuminating potential pathways for the creation of more proficient combined monitors.

\section{Which recovery mechanisms can be used to build safety monitors?} \label{sec:recovery_actions}

In Section~\ref{sec:detection}, we explored a plethora of methods aimed at detecting unsafe predictions stemming from ML-based perception functions. 
When activated, these detection mechanisms raise a safety alert to the autonomous system, necessitating immediate and appropriate measures to avoid hazardous situations. 
In most ML monitoring studies, the basic alert is a basic flag, and more complex actions are usually not investigated.
We call such actions \emph{recovery mechanisms}, and this section is dedicated to their detailed examination.  

\subsection{Switching the control system}
The most straightforward and widely used approach when detecting a potentially dangerous error is to switch to a simpler control algorithm, not relying on the faulty ML perception component. Phan \etal\cite{phan2019neural} proposed the neural simplex architecture, which switches from a high-performance ML-based controller to a simpler safe controller when unsafe behavior is detected. Adapting this architecture to ML-based perception functions would be highly valuable but challenging. Indeed, in many practical scenarios, it is hard to design simple controllers that do not rely on ML for state estimation. As a result, for complex autonomous systems relying on visual perception, the default recovery actions often consist of switching back control to a human driver\cite{stocco2020misbehaviour} or triggering an emergency braking.\cite{cofer2020run} The former can be hazardous when a fast reaction is required. The latter might not be safe for specific scenarios, e.g., autonomous cars driving on the highway. We believe that the challenging task of coming up with control procedures to bring complex autonomous systems back to safety is of significant importance and should receive more attention.

\subsection{Immediate prediction enhancement}
For some specific threats, adapted safety measures can be used to improve the predictions of the ML-based perception function and increase confidence in the system.

\subsubsection{Input reconstruction}
A family of approaches consists of improving the quality of the ML inputs. These techniques are used when one can identify the cause of the wrong prediction and mostly consist of removing the specific type of noise. Recently, most image-denoising approaches use autoencoders, trained to reconstruct the original image without the identified noise pattern.\cite{gondara2016medical} For example, image dehazing algorithms can be used to react to fog or smoke.\cite{abdulkareem2021new} Other approaches have been proposed to react to different light exposure conditions,\cite{yan2022high, yan2021dual} saturation,\cite{liu2021adnet} water drops,\cite{liao2013video, qian2018attentive} or even more standard Gaussian noise or impulse noise.\cite{zhou2020awgn,zhang2021countering} 
Other works proposed to enhance the original image by increasing its resolution artificially\cite{wang2020deep} by using convolutional neural networks\cite{shi2016real} and generative adversarial networks.\cite{ledig2017photo,marinescu2021bayesian} Furthermore, to deal with images having chunks of pixels damaged by sensor failures, image inpainting techniques can be used.\cite{saharia2021palette} Finally, to handle errors related to incomplete color information (mosaic-styled images), some works have applied demosaicing techniques\cite{ni2020color} or gradient-based feature extraction.\cite{zhou2021gradient}

\subsubsection{Changing final prediction}
To respond to the detection of an adversarial example, Al-Afandi and Horvath\cite{alclass} proposed to exclude the predicted classes (corresponding to the attack) and to study the resulting loss landscape to recover the original class. On the other hand, Li \etal\cite{li2021reversibility} proposed an approach to reverse the effect of an adversarial attack on a classifier by studying the behavior of adversarial examples and establishing a mapping between true classes and predicted classes. Whether similar approaches can be used to respond to different kinds of threats is still an open question that is worth investigating. For example, the work by Salay \etal\cite{salay2019safety} could be extended to downgrade classification at runtime, e.g., if a high classification uncertainty is detected between the classes ``car'' and ``truck'', the prediction could be changed to the sup-class ``vehicle''.

\subsubsection{Using alternative components}
When an error is detected in the state estimation, a possible approach is to substitute some components of the perception pipeline and recompute its output. For example, when a sensor failure is causing the perception error, one can rely on existing backup sensors to still compute the expected prediction.\cite{kakamanshadi2015survey} When high uncertainty is detected, one might use sensors with higher resolution locally (spatially or temporally) to get a better prediction.\cite{guerin2022_icra} However, these high-resolution sensors might not be usable in the regular operation pipeline because of processing time or energy consumption constraints. Another idea is to use an ensemble of ML models with different coverage mechanisms to replace unsafe predictions.\cite{gontijo2021no}

\subsection{Impact on long-term system enhancement}
Finally, we also discuss how runtime threats identified by a monitoring detection mechanism can be used to improve the system's safety in the long run. A common strategy in the industry is to store a huge amount of frames while the system is running for posterior offline labeling and model retraining.\cite{tesla} Post-labeling is usually done manually, using other sensors, or automatically using other ML models. Specific continual-learning approaches exist for model retraining, particularly to avoid catastrophic forgetting when incorporating novel classes.\cite{van2019three,liu2021adaptive} If one chooses to collect data detected as unsafe by a monitoring system for retraining, it will have a positive long-term impact on the safety of the system.

Although these approaches do not guarantee the immediate safety of the system, they are essential to reduce safety-critical errors in the long run. In addition, storing data detected as threats can also be useful for building relevant datasets to test future developments of safety-critical systems.

\section{How to evaluate safety monitors?} \label{sec:evaluation}

A proper evaluation protocol for safety monitoring should ensure that the Safety Monitor (SM) presents the following characteristics:
\begin{enumerate}
    \item It guarantees that the system never reaches any safety-critical state.
    \item It maintains a high availability of the system.
    \item It complies with the runtime constraints of the system (execution time, hardware capacity).
\end{enumerate}
This section delves into the evaluation practices adopted for ML safety monitoring. We first examine the assessment methods for detection mechanisms. Being generic modules, they often permit independent evaluations detached from the complete system. Then, we explore the existing evaluation protocols used for approaches that monitor ML components at the system level. Finally, we discuss how the impact of research on SM could be increased by standardizing evaluation methodologies. 

\subsection{Evaluation of detection mechanisms}
Most detection approaches intend to identify unsafe input data, which are likely to produce erroneous output when processed by the ML model. Ensuring the efficacy and safety of these methods necessitates the crafting of pertinent test datasets, coupled with the application of appropriate evaluation metrics. This section is dedicated to elucidating these crucial aspects.

\subsubsection{Evaluation datasets}
In this section, we present how datasets are built to evaluate safety monitoring detection mechanisms.
This process can be seen as fault injection, focusing on the threats presented in Section~\ref{sec:threats}.

\paragraph{Novelty detection}
Novelty detection aims at identifying when the label of an input does not belong to any of the predefined classes handled by the ML model. Thus, to evaluate the capacity of an SM to identify novel inputs, most approaches have used modified versions of standard image classification datasets such as Imagenet,\cite{deng2009imagenet} MNIST\cite{deng2012mnist} or CIFAR.\cite{krizhevsky2009learning} In particular, two main strategies can be used to create novelty detection datasets. The first one consists of merging two datasets with non-overlapping class labels. One dataset is used to fit the model (training split), and its test split represents the in-distribution data for evaluation, while the other dataset serves as novelty data. This approach was used for the experiments of many papers on out-of-distribution detection.\cite{sun2021react, schorn2020facer, lee2018simple, hendrycks2016baseline, liang2018enhancing, hsu2020generalized, shafaei2018less, ferreira2021benchmarking} The second strategy splits a single dataset into two subsets with disjoint labels.\cite{denouden2018improving, sabokrou2018adversarially, lukina2020into, henzinger2020outside, ferreira2023sena, wu2021customizable, roitberg2018informed} Recently, Wang \etal\cite{wang2022vim} attempted to build a less ambiguous novelty benchmark dataset by asking two independent human annotators to label novelty images for ImageNet.

\paragraph{Distributional shift detection}
A distributional shift occurs when the marginal distribution of the runtime input data differs from the training distribution, while the label set does not change. It can come from sensor failures, changes in external conditions, or modifications to the environment itself. Thus, most papers have relied on injecting perturbations to test images to evaluate runtime detection mechanisms for detecting such shifted data. Several papers have proposed to inject artificial corruption\cite{hendrycks2019benchmarking} into standard image classification datasets.\cite{rahman2019did, schorn2020facer, hendrycks2016baseline, liang2018enhancing, hsu2020generalized, ferreira2021benchmarking} Others have injected faults into realistic autonomous systems scenarios. Some have used autonomous vehicle simulators, such as CARLA,\cite{Dosovitskiy17} to simulate different kinds of driving conditions (weather, light).\cite{feng2021robust, cai2020real, ferreira2022simood} Others have applied artificial perturbations to existing real-world datasets for autonomous driving\cite{chen2020task, chen2021monitoring, zhang2018deeproad, zhou2019automated} or UAV emergency landing.\cite{guerin2022_icra}

\paragraph{Adversarial detection}
This setting represents an intentional modification of in-distribution data to make a deep learning model fail. The main approach to evaluate an SM detector at this task consists of applying an adversarial attack (see Section~\ref{sec:adversarial} for examples) to the test dataset under evaluation to constitute a binary classification dataset containing both normal and attacked images. This has been applied to standard image datasets\cite{kantaros2020visionguard, wang2019adversarial, liu2020input, ferreira2021benchmarking} and simulated autonomous driving scenarios.\cite{cai2020real, chen2020task}

\subsubsection{Evaluation metrics}

In Section~\ref{sec:threats}, we highlighted two competing perspectives in the domain of unsafe ML prediction detection. One view centers on out-of-distribution (OOD) detection, which seeks to recognize specific runtime threats, assuming that all standard data should be embraced, while any altered data should be excluded. In contrast, the out-of-model-scope (OMS) detection focuses on pinpointing the ML model's actual prediction errors. This latter perspective has been adopted in several research works.\cite{cheng2019runtime, wang2020dissector, hendrycks2016baseline, ferreira2021benchmarking, kang2018model, granese2021doctor}. In our recent work,\cite{guerin2023outofdistribution} we argue for broader adoption of OMS detection, postulating that the concept of ``OODness'' is ambiguous and that in-distribution errors should also be addressed by competent monitors. Despite these distinct paradigms, the evaluation datasets remain consistent across both views.

Detection mechanisms are binary classifiers, hence, switching between these perspectives boils down to toggling the binary monitoring labels. Then, the performance of a monitoring approach is gauged through standard binary classification metrics. In this work, we consider that a True Positive (TP) is a rejected invalid input, while a True Negative (TN) is an accepted valid input. A quick rundown of binary classification metrics includes:
\begin{itemize}
    
\item \textbf{Accuracy --} Proportion of correctly classified inputs. It can be misleading when the dataset is not balanced.

\item \textbf{FP rate --} Proportion of valid inputs that were rejected.
    
\item \textbf{FN rate --} Proportion of invalid inputs that were missed.
    
\item \textbf{TPR@95TNR --} TP Rate when the TN Rate is $0.95$. It represents the probability of finding invalid data when the rejection threshold is set so that $95\%$ of valid data are accepted. 
    
\item \textbf{AUROC --} Area Under the Receiver Operating Characteristic (TPR against FPR). AUROC is threshold-independent and represents the probability that rejection scores of valid inputs are lower than invalid ones.
    
\item \textbf{Precision --} Proportion of rejected inputs that were invalid.
    
\item \textbf{Recall --} Proportion of invalid inputs that were rejected.
    
\item \textbf{AUPR --} Area Under the Precision-Recall curve. AUPR is better than AUROC when the positive class and negative class have greatly differing base rates.

\item \textbf{P@80R --} Precision when the recall is set to 0.8.\cite{rahman2019did}

\item \textbf{F1-score --} Harmonic mean of the precision and recall. This score represents a unified performance evaluation when the rejection threshold has been fixed.
    
\item \textbf{Matthews Correlation Coefficient --} It accounts for all categories of the confusion matrix (TP, FP, TN, FN).

\end{itemize}

When the monitored model addresses a different task than classification (e.g., regression), the definition of prediction failure is not as straightforward. For example, a neural network predicting the steering angle of a vehicle for the next time step will always commit some degree of error, and defining failure requires choosing a threshold for these errors. For such cases, to assess the performance of a detection mechanism, one can compare the values of task-specific metrics between accepted and rejected images. Examples of such metrics include average precision for object detection, mean squared error for regression tasks, and intersection over union for semantic segmentation.

Furthermore, it's crucial to account for the detection mechanism's computational performance in terms of execution time and memory usage to gauge the overall system overhead when integrating such safety monitors.

\subsection{System-level evaluation}
When an ML-based perception component is embedded into the control loop of an autonomous system, not all prediction errors will lead to the same outcomes. For example, some perception errors can generate catastrophic events (e.g., missing a pedestrian crossing a road)\cite{salay2019safety}. In contrast, others might not even change the system's behavior (e.g., detecting a tree as a street lamp). In addition, Haq \etal\cite{haq2020comparing} showed that offline testing (unit tests) is
more optimistic than online testing (simulation) since several safety violations that were identified in simulation could not be identified offline.
For this reason, it is important to evaluate how well a safety monitor is performing within the context of the system in which it is integrated. 

Recent works have designed safety monitors in a real-world application context, where the impact of a prediction by the perception component can be assessed. For such cases, it is thus possible to evaluate different aspects of the performance of an SM, such as the added safety and the loss of system availability. An example was proposed by Stocco et al.,\cite{stocco2020misbehaviour, stocco2020towards} where the monitor is implemented in a simulation environment for an end-to-end autonomous driving scenario. This way, they can play the same scenarios with and without the monitor and evaluate when critical misbehavior has been avoided (added safety) and when interventions were unnecessary (loss of availability). Another simulation-based SM evaluation was conducted by Cofer \etal\cite{cofer2020run} for an aircraft taxiing application. On their experimental dataset, they were able to avoid all the cases where the neural network led the aircraft to exit the runway thanks to their safety monitoring system. On another note, Guerin \etal\cite{guerin2022_icra} evaluated safety monitors for a drone emergency landing scenario. By defining a safety score for any landing zone, they can compare the emergency landing system with and without the monitors. Different monitors can be compared based on their safety benefits to the system. Guerin et al.\cite{guerin2022issre} endeavored to provide a theoretical foundation for evaluating ML monitors within autonomous systems. They delineated three metrics: Safety Gain, Residual Hazard, and Availability Loss, and demonstrated their computation across various examples of ML-enabled systems. 

\subsection{Evaluation coverage}
When performing evaluation, we need to consider two different scenarios: simulated, and real-world scenarios. Each presents distinct advantages and challenges.

To test perception functions, it is frequent to use a simulation environment.\cite{stocco2020misbehaviour, loquercio2020general, buerkle2021fault} This allows to reset the environment to a previous configuration and compare the responses to different perception and monitoring outputs. However, the existence of a reality/simulation gap is frequent. Indeed, there are usually significant differences between simulation environments and real-world scenarios, where new, unexpected threats can happen.\cite{zandbergenevaluating} On the other hand, when evaluating a perception function on real images, the collected data never represent all possible threats to which a safety-critical system might be exposed. Nevertheless, an exhaustive safety analysis of the system might help cover a higher proportion of these threats.\cite{borg2019safely}
     
Even when evaluations are conducted in representative test scenarios, it is hard to evaluate the performance of a perception function, and safety monitor as the ground truth is often not available at runtime. This is often referred to as the oracle problem.\cite{jahangirova2017oracle} As a result, all frames and sensor values must be recorded for off-line labeling and performance and safety evaluation. 
Such evaluations need to be done periodically to avoid a decrease in ML performance and system safety due to dataset shifts.\cite{rabanser2019failing}

\subsection{Evaluation for certification}
Specific evaluation and certification procedures for autonomous systems were proposed in the literature. Myers and Saigol\cite{myers2020pass} developed a framework to assess the safety of autonomous driving by applying two types of outcome-scoring rules: prescriptive and risk-based. The first contains measurable rules, which must always be verified, while the second contains undesirable outcomes that must not occur too often. De Gelder and Den Camp\cite{de2020procedure} proposed a certification scenario for self-driving vehicles, considering three stakeholders: the applicant, the assessor, and the road/vehicle authority. The applicant applies for the approval of one specific autonomous vehicle. The assessor assesses this vehicle and advises the authority, who sets the requirements and approves the vehicle for road testing. De Gelder \etal\cite{de2021risk} proposed a risk analysis expressed as the expected number of injuries in a potential collision to compare it to road crash statistics. The authors decompose the quantified risk into the three aspects stipulated by the ISO-26262 and ISO/DIS-21448 standards: exposure, severity, and controllability. On another note, Guerin \etal\cite{guerin2021_ssiv} assessed the requirements to certify UAV operations in urban environments using a document called Specific Operations Risk Assessment (SORA),\cite{SORA} which provides guidelines to develop and certify safe UAV operations.

\section{Conclusion and open challenges}\label{sec:conclu}

Machine Learning solutions are being used increasingly to build perception functions for autonomous systems, but they cannot be trusted for safety-critical applications.
Safety Monitors aim to ensure that the system always remains in a safe state despite the occurrence of faults.
This work presents a comprehensive survey about safety monitoring of ML perception functions, addressing every step of the development process, i.e., threat identification, requirements elicitation, detection of failure and reaction, and evaluation.
We present existing works related to SM and highlight the current gaps in the literature to reach the level of integrity required for such safety-critical systems.
After conducting this extensive study, we consider that the field's biggest limitations and open challenges are the following.

\textbf{Defining safety monitoring objectives --}
More research should be conducted regarding how to formulate the monitoring requirements to reflect the outcomes of
the safety analysis and other relevant properties of the system. To illustrate this, we can mention a result from Ferreira \etal,\cite{ferreira2021benchmarking} which showed that most detection mechanisms based on out-of-distribution detection (threat identification) suffer from a high number of false positives and false negatives when considering their ability to detect a failure of the ML model. This limitation results from a misalignment between SM specifications and system-level objectives. Indeed, not all threats lead to errors, and some in-distribution images lead to wrong predictions.

\textbf{Choosing detection and reaction mechanisms -- }
Different types of detection and reaction mechanisms were presented in this work. Such approaches must be properly combined with the task at hand to build a good safety monitor, which is difficult due to the vast possibilities.
This choice is highly dependent on the application context, but we believe that meaningful research could be proposed to map task characteristics to detection/reaction mechanisms.
For example, identifying that certain generic detection mechanisms are suitable for specific kinds of threats would be useful. Likewise, it would be valuable to study whether specific recovery actions can improve the performance of the ML model when combined with specific detection mechanisms.

\textbf{Combining safety monitoring architectures -- }
It is probably desirable to use several monitoring approaches for different aspects of the perception task.
For example, specific detection mechanisms could be responsible for different regions of the input space or different types of threats.
Additionally, strategies that are not purely based on data, such as plausibility checks,\cite{kontos2021phase}
model assertion,\cite{kang2020model} and classification failure mode and effects analysis,\cite{salay2019safety}
can be applied to complement data-based monitors.
Studying how to combine several safety monitors and verify the consistency of their outputs is an important open challenge
for the field.

\textbf{Implementation constraints -- }
The safety monitors discussed in this work are expected to perform within embedded systems.
Hence, it is essential to design SMs that can function under limited computing power and memory.
Additionally, they must adhere to the system's energy consumption constraints.
Being executed at runtime, SMs also need to meet stringent execution time criteria to ensure seamless synchronization with the primary system being monitored.

\textbf{Standardized evaluation -- } 
As explained earlier, many different test datasets and evaluation metrics are used by practitioners who select evaluation procedures based on their own needs and use cases. Such evaluation scenarios might differ between domains (e.g., automotive, avionics, naval), making it difficult to compare different safety monitoring approaches. 
We believe that the development of a unified benchmarking framework, including different autonomous system use cases and evaluation metrics, is a promising research direction. Indeed, it will foster the development of safety monitoring approaches that will help certify safety-critical systems that rely on ML models. 

\textbf{Certification -- } 
With the increasing use of ML approaches, solutions such as safety monitors will be an important tool to certify future critical autonomous systems.
Hence, the community needs to start addressing the challenges mentioned above.
As a first step, we believe that building unified evaluation benchmarks and metrics, reflecting the different aspects highlighted in this survey
would greatly help to develop SM better suited to the safety-critical context. To go further and be able to certify perception components and their safety monitors, one needs to establish the impact of perception errors on the safety of the entire system, which is an unsolved problem. 

\bmsection*{Acknowledgments}
This research has received funding from the European Union’s Horizon 2020 research and innovation program under the Marie Skłodowska-Curie grant agreement No 812.788 (MSCA-ETN SAS). This publication reflects only the authors’ view, exempting the European Union from any liability. Project website: http://etn-sas.eu/.

This research has also benefited from the AI Interdisciplinary Institute ANITI. ANITI is funded by the French ”Investing for the Future – PIA3” program
under the Grant agreement No ANR-19-PI3A-0004.

\end{document}